%% file: main.tex
\documentclass{neus2025}

\title[Reward Shaping and Action Masking for Compositional Tasks using BTs and LLMs]{Reward Shaping and Action Masking  for Compositional Tasks using Behavior Trees and LLMs}

\usepackage{amsmath,amssymb,amsfonts}
\usepackage{algorithm}
\usepackage{algpseudocode}
\usepackage{graphicx}
\usepackage{caption}
\usepackage{subcaption}
\usepackage{textcomp}
\usepackage{array}
\usepackage{multirow}
\usepackage{appendix}
\usepackage{float}
\usepackage{makecell}
\usepackage{booktabs}
\usepackage{wrapfig}

\AtBeginDocument{%
  }
\coltauthor{\Name{Nicholas Potteiger} \Email{nicholas.potteiger@vanderbilt.edu}\\
 \Name{Ankita Samaddar} \Email{ankita.samaddar@vanderbilt.edu}\\
 \Name{Taylor T. Johnson} \Email{taylor.johnson@vanderbilt.edu}\\
 \Name{Xenofon Koutsoukos} \Email{Xenofon.Koutsoukos@vanderbilt.edu}\\
 \addr Vanderbilt University, Nashville, TN}

\begin{document}

\newtheorem{exmp}{Example}

\maketitle

\input{chapters_v2/abstract}


\input{chapters_v2/introduction}
\input{chapters_v2/formulation}
\input{chapters_v2/btasreward}

\input{chapters_v2/evaluation}
\input{chapters_v2/advantages}
\input{chapters_v2/conclusion}

\newpage
\bibliography{ref}

\newpage
\appendix

\input{chapters_v2/background_related}
\newpage
\input{appendix/nesy_rl_loop}
\input{appendix/logic_specifications}
\input{appendix/env_task_descriptions}
\input{appendix/experiment_setup}
\newpage
\input{appendix/verification_times}
\newpage
\input{appendix/exp_results}
\input{appendix/hrm}
\input{appendix/using_demonstrations}

\end{document}

%% file: chapters_v2/abstract.tex
\begin{abstract}
    Decomposing complex tasks into a sequence of simpler subtasks can improve learning efficiency for an autonomous agent. Reinforcement learning (RL) can be used to optimize agent policies to complete subtasks, but requires well-defined subtask rewards and benefits from action masking. Recent work uses large language models (LLMs) to automate reward shaping and action masking, however none of them fully address reactivity to subtask failure and modularity to varying objects for compositional tasks. To overcome these challenges, we develop masking reward behavior tree (MRBT), a symbolic structure used as a reactive and modular reward and action mask function. We design an MRBT template and derive logical specifications to construct and verify MRBTs for a sequence of object-interaction subtasks.  Further, we develop an automated pipeline that uses an LLM to generate MRBTs robust to varying task objects, an SMT-solver to verify correctness of specifications, and a neurosymbolic RL loop to train agents on compositional tasks.
    Experiments demonstrate successful generation and refinement of five MRBTs, consistently improving training efficiency and task success rates over baselines and MRBTs without action masking. 
    We further highlight three advantages of MRBTs: transferability, modularity, and verifiability. 
\end{abstract}

%% file: chapters_v2/introduction.tex
\section{Introduction}
Decomposing complex tasks allows autonomous agents to learn simpler behaviors. Many robotic tasks, such as assembly and drone delivery, involve sequential subtasks requiring navigation and manipulation of objects. Although reinforcement learning (RL) can train policies for these subtasks, it benefits from reward shaping and action masking to improve learning efficiency (\cite{huang2020closer}). Designing such rewards and action masks manually is challenging, as it requires identifying subtasks, ensuring reactivity to subtask failures, and maintaining modularity across varying task objects. These challenges motivate automated design of reactive, modular rewards and action masks for compositional tasks.

To automatically design rewards and action masks, previous work has leveraged large language models (LLMs) or vision language models (VLMs). They can be fine-tuned to evaluate task completion (\cite{du2023flamingosuccessdetector, fan2022minedojo}). Other approaches use pre-trained models to generate reward code (\cite{ma2023eureka, venuto2024codeasreward, Qu2025latentrewardllm, qi2025roboticskillrewardsllms}) or construct reward machines (\cite{alsadat2025llmsrm, alsadat2025llmswarmrm}). Similar to our work, VLM-CaR (\cite{venuto2024codeasreward}) decomposes tasks into a sequence of subtask reward code that is tested using expert demonstrations. Very few works have examined how  LLMs can be used for action masking, with one work using LLM outputs to generate dynamic action masks (\cite{zhao2025camelcontinuousactionmasking}). While prior work provides a strong foundation for automatic reward shaping and action masking, it does not fully address reactivity or modularity in compositional tasks, nor does it consider joint reward and action mask functions. Related work section is in Appendix~\ref{app: related_work}.

To address reactivity and modularity, we develop a reward and action mask function using a behavior tree (BT). BTs are interpretable, reactive, and modular, making them well-suited for compositional tasks and LLM reasoning. The BT, referred as masking reward BT (MRBT), outputs rewards and action masks from leaf behaviors while the BT coordinates execution. We design an MRBT template for sequential object-interaction subtasks, with placeholders for subtasks, action masks, and first-order logic formulas conditioning rewards and action masks. The template enables reactive backtracking for subtask failure. We derive logic specifications to verify the correctness of the logic formulas with respect to task completion. The specifications are verified over finite-horizon trajectories generated by an SMT solver under a symbolic encoding of the environment’s transition dynamics.

To automatically design rewards and action masks that are reactive to subtask failure and modular to varying task objects, we leverage the MRBT template, an LLM, and an SMT solver. We define a task space that encapsulates possible combinations of the same task structure with varying task objects. The LLM inputs the MRBT template and task-specific information to generate MRBTs that are robust to any task in a task space. The SMT-solver checks satisfiability of the specifications, re-prompting the LLM if unsatisfiable. The MRBT is integrated into a neurosymbolic RL loop to train an agent over the task space.

% technical contributions
The main technical contributions of this work are:

\noindent
1) Reward and action mask functions for compositional tasks using \textit{masking reward behavior tree (MRBT)}. An MRBT outputs rewards and action masks from executed leaf behaviors.

\noindent
2) An MRBT template to construct MRBTs for sequential object-interaction subtasks. In addition, logical specifications are derived to check subtask completion, proximity to subtask objects prior to subtask completion, and that subtasks can be composed to yield maximal reward without regression due to subtask failure. They are verified with an SMT-solver constrained by a symbolic model of the environment dynamics. Alternatively, expert demonstrations can be used for testing specifications.

\noindent
3) An automated pipeline\footnote{Code and LLM prompts available at \url{https://github.com/npotteig/bt_as_reward}} for generating and verifying MRBTs from the MRBT template that are robust to any task in a task space. The MRBTs are integrated into a neurosymbolic RL training loop for optimizing an agent to complete any task in a task space. 

\noindent
4) A comprehensive evaluation of MRBTs is conducted, including pipeline analysis and an ablation study against two baselines across five task spaces in two environments. ChatGPT-5 (\cite{openai2025gpt5}) and Z3 (\cite{de2008z3}) are used to generate, verify, and refine MRBTs. Higher training efficiency and task success are consistently achieved using an MRBT compared to baselines and an MRBT without action masking. In the most complex task space, $\geq 80\%$ average success rate is achieved, compared to $\leq 70\%$ for baselines.

\noindent
5) An evaluation of advantages of MRBTs. Agent policies trained using MRBTs transfer navigation skills to a realistic quadcopter simulator (AirSim (\cite{airsim2017fsr})), exhibit greater modularity than reward machines as task complexity grows, and offer theoretical guarantees via Z3.

%% file: chapters_v2/formulation.tex
\section{Autonomous Agents for Compositional Tasks}

We study compositional tasks expressed in natural language as sequences of subtasks with object-interaction, where a template defines a task space of structurally equivalent instances that vary only in the task objects. 
We represent a task space as $\mathcal{M} = \langle \ell,  \mathcal{V}, \mathcal{G} \rangle$, where $\ell$ is a natural language template, $\mathcal{V}$ is a set of discrete symbolic variables that fill the template, and $\mathcal{G}$ is a set of first-order logic formulas that represent task completion.

The agent must complete tasks in an environment $\mathcal{E}$ represented as a partially observable Markov decision process with logic formulas, $\mathcal{E} = \langle \mathcal{S}, \mathcal{O}, \mathcal{A}, \allowbreak \omega, p, \mathcal{L}, l, r, \gamma \rangle$. $\mathcal{E}$ consists of a state space $\mathcal{S}$, an observation space $\mathcal{O}$, an action space $\mathcal{A}$, an observation function $\omega: \mathcal{S} \times \mathcal{A} \rightarrow \mathcal{O}$, a transition function $p: \mathcal{S} \times \mathcal{A} \rightarrow \mathcal{S}$, a finite set of first-order logic formulas $\mathcal{L}$ constructed from predicates derived from $\mathcal{S}$, a labeling function $l: \mathcal{M} \times \mathcal{S} \rightarrow 2^\mathcal{L}$ that maps tasks and states to subsets of logic formulas we refer to as labels, a reward function $r: (2^\mathcal{L})^\text{+} \rightarrow \mathbb{R}$, and discount factor $\gamma$. The reward function inputs a label history $\lambda_{t+1} = \langle l(s_0, m), ..., l(s_t,  m), l(s_{t+1}, m) \rangle \in (2^\mathcal{L})^+$ to compute a scalar reward.
Further, an action mask function $\mu:(2^\mathcal{L})^\text{+} \rightarrow 2^\mathcal{A}$ can dynamically restrict available actions. The task input to the labeling function are values in $\mathcal{V}$ and $\mathcal{G}$.

% aim is to find a policy that completes every mission in the misison space
The objective of the agent is to find a policy $\pi^*: \mathcal{M} \times \mathcal{O} \rightarrow \mathcal{A}$, where for every task in a task space $\forall m^i = \langle \ell, v^i, g^i \rangle \in \mathcal{M}$, $\pi^*$ generates a trajectory of observation-action pairs $\tau^i = ((o_0, a_0), ..., (o_T, a_T))$ where the terminal state $s_T$ satisfies $g^i$. The task input is a sequence of integers, where each integer represents a word in the natural language task.

% Input
The problem is to use an LLM to design reward and action mask functions that guide the agent toward learning $\pi^*$. Given a task space $\mathcal{M}$ and environment $\mathcal{E}$, the LLM takes as input task-specific information such as, $\mathcal{M}$, predicates from $\mathcal{S}$, and $\mathcal{A}$.
% Output
The output is a set of logic formulas $\mathcal{L}$, labeling function $l(s, m)$, reward function $r(\lambda)$, and action mask function $\mu(\lambda)$, given label history $\lambda \in (2^\mathcal{L})^+$, state $s \in \mathcal{S}$, $\forall m \in \mathcal{M}$. 
% These functions should be successful in training the agent using RL to complete any task in the task space.

We use MiniGrid LockedRoom (\cite{MinigridMiniworld23}) as a running example, where the agent must find a key in one of six rooms, unlock a door, and reach the goal. The task space is defined by 
$\ell =$ \textit{``get the \{key\_color\} key from the \{room\_color\} room, unlock the \{door\_color\} door and go to the goal''}, $\mathcal{V} = \{\textit{key\_color}, \textit{room\_color}, \textit{door\_color}\}$, and $\mathcal{G}$ contains a single logic formula $g^0$ that checks when the agent reaches the goal. Color values are drawn from \textit{\{red, green, blue, purple, yellow, grey\}}.
The state space $\mathcal{S}$ is an $N \times N$ grid with object information, including the agent, keys, doors, and the goal. Predicates derived from $\mathcal{S}$ include $agent\_pos$, $key\_pos$, $door\_pos$, $door\_state$, and $goal\_pos$, where positions are 2D tuples and $door\_state \in \{\textit{open}, \textit{closed}, \textit{locked}\}$. Object attributes may be indexed by color (e.g. $key\_pos_{red}$), and missing or occluded objects are assigned a value of $-1$ for position and state. The action space is
$\mathcal{A} = \{\textit{left}, \textit{right}, \textit{forward}, \textit{pickup}, \textit{drop}, \textit{toggle}, \textit{done}\}$.

% We use MiniGrid LockedRoom~\cite{MinigridMiniworld23} as a running example; the objective is for the agent to find a key in one of six rooms and then unlock a door to get to the goal. The task space for LockedRoom is defined as $\ell =$ \textit{``get the \{key\_color\} key from the \{room\_color\} room, unlock the \{door\_color\} door and go to the goal"}, $\mathcal{V} = \textit{\{key\_color, room\_color, door\_color\}}$, $\mathcal{G}$ contains one logic formula $g^0$ that evaluates to \emph{True} if the agent is at the goal position. The color values are \textit{"red", "green", "blue", "purple", "yellow", "grey"}. The state space $\mathcal{S}$ is a $N \times N$ grid of cells with object information. Objects include the agent, keys, doors, the goal, and more. Predicates derived from $\mathcal{S}$ include variables such as $agent\_pos$, $key\_pos$, $door\_pos$, $door\_state$ and $goal\_pos$. Positions are two dimensional tuples and $door\_state$ can be one of "open", "closed", or "locked". $key\_pos$, $door\_pos$, $door\_state$ can be indexed by color (e.g. $key\_pos_{red}$). If an object is occluded or not present its position and state are $-1$. The action space $\mathcal{A}$ is \textit{\{left, \allowbreak right, \allowbreak forward, pickup, drop, toggle, done\}}. 

%% file: chapters_v2/btasreward.tex
\begin{figure}[t]
    \centering
    \includegraphics[width=0.8\linewidth]{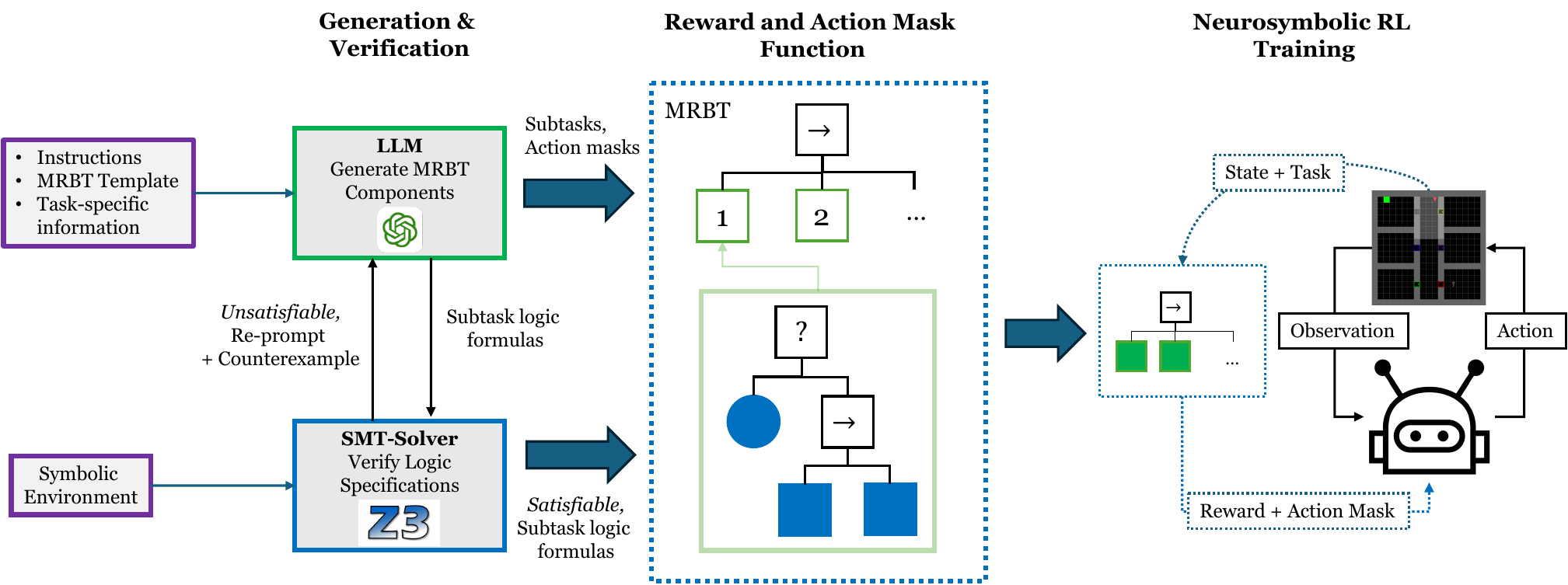}
    \caption{Automated pipeline to \textit{generate} and \textit{verify} MRBTs and their use in \textit{training}.}
    \label{fig:approach_flow}
    \vspace{-0.5cm}
\end{figure}

\section{Masking Reward Behavior Trees}

To learn agent policies for compositional tasks, we define a masking reward behavior tree (MRBT) used as a reactive and modular reward and action mask function. We design a template to construct MRBTs for sequences of object-interaction subtasks. From these MRBTs, we derive logical specifications to verify correctness of MRBT logic. Finally, we present an automated pipeline (Figure~\ref{fig:approach_flow}) that uses the template, specifications, an LLM, and an SMT solver to generate and verify MRBTs, which are then integrated into a neurosymbolic RL loop.

\subsection{Preliminaries}
% Behaviour trees
%\subsubsection
\vspace{-0.1cm}
\noindent
\textbf{Behavior Trees:} BTs are symbolic policies known for reactivity, modularity, and interpretability. A BT executes by propagating a tick starting from its root through a finite set of behaviors in depth-first order; each behavior returns \textit{Success}, \textit{Running}, or \textit{Failure}. Internal nodes are control nodes, such as Sequence and Fallback that control propagation of the tick to its children. Leafs nodes are execution nodes, such as Condition and Action, that execute behavior in the environment when ticked.

% Reward Machines
%\subsubsection
\noindent
\textbf{Reward Machines: }\label{sec: rew_mach}
% Regular definition
% Simple Reward Machine
A reward machine (RM) is a finite state machine that takes abstracted descriptions of the environment as input and outputs reward functions (\cite{icarte2018rewardmachine}). Given a set of propositions, or more generally first-order logic formulas, $\mathcal{L}$, state space $\mathcal{S}$, and action space $\mathcal{A}$, an RM is a tuple $\langle U, u_0, F, \delta_u, \delta_r \rangle$ where $U$ is a finite set of machine states, $u_0 \in U$ is the initial state, $F$ is a finite set of terminal machine states such that $U \cap F = \emptyset$, $\delta_u: U \times 2^\mathcal{L} \rightarrow U$ is the state-transition function, $\delta_r: U \times 2^\mathcal{L} \rightarrow \big[ \mathcal{S} \times \mathcal{A} \times \mathcal{S} \rightarrow \mathcal{S} \big]$ is a state-reward function that outputs a reward function based on the RM state. A \textit{simple reward machine} is only dependent on $\mathcal{L}$ where $\delta_r: U \times 2^\mathcal{L} \rightarrow \mathbb{R}$ outputs a scalar reward. 

%\subsubsection
\noindent
\textbf{Neurosymbolic Reinforcement Learning:} Neurosymbolic RL is an emerging class of methods that aims to enhance reasoning and learning by combining symbolic components (e.g., BTs) with RL components (e.g., neural networks) (\cite{acharya2024nesyrl}, \cite{design2024nick}). The methods fall under one of three categories: \textit{learning for reasoning}, \textit{reasoning for learning}, and \textit{learning-reasoning}. RMs fall under reasoning for learning, as they serve as symbolic components to guide the output of a neural policy by reward shaping. Our work falls under reasoning for learning. We use a BT to provide feedback to the agent.

\subsection{Masking Reward BT Formalization}
A masking reward BT (MRBT) is a type of BT that inputs abstracted descriptions of the environment and outputs a real-valued reward and discrete action mask. The execution nodes (leaves) of an MRBT are simple RMs with action masks defined as follows:

\begin{figure*}[t]
    \centering
    \includegraphics[width=0.7\linewidth]{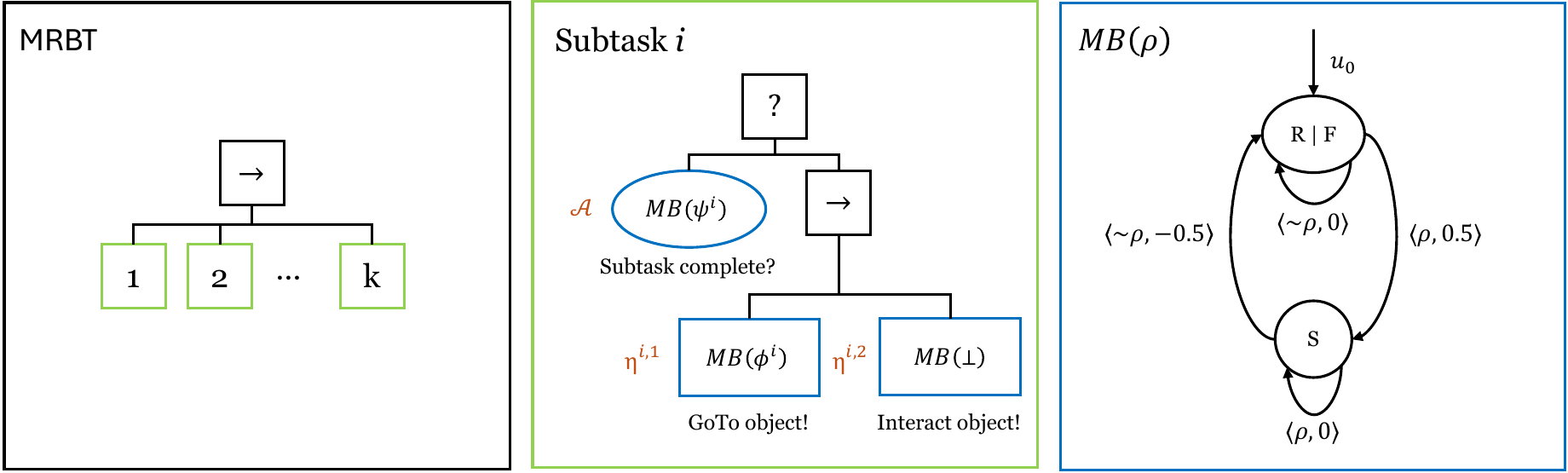}
    \caption{Masking reward behavior tree (MRBT) template.}
    \label{fig:rbt_template}
    \vspace{-0.5cm}
\end{figure*}

% masking behavior RMs
\begin{definition}
    (masking behavior reward machine) Given a set of $\mathcal{L}$ logic formulas and discrete action space $\mathcal{A}$, a masking behavior reward machine is $MB = \langle U, u_o,  \delta_u, \delta_r, \eta \rangle$, where $U$ is a set of states $\{Success, Running, Failure\}$, $u_0$, $\delta_u$, and $\delta_r$ are defined as in a simple reward machine (see Section~\ref{sec: rew_mach}), and $\eta \in  2^\mathcal{A}$ is an action mask. 
\end{definition}

Masking behavior reward machines (MBRMs) are structured into an MRBT according to the following definition:

% An MRBT transitions between states in its masking behavior RMs (MBRMs) and emits rewards and action masks according to the following definition:

\begin{definition}\label{def:mrbt}
    (masking reward behavior tree) An MRBT is a BT $\mathcal{T}$ whose leaves are a set of MBRMs $\mathcal{B}_{leaf}$ defined over a common set of logic formulas $\mathcal{L}$ and action space $\mathcal{A}$. An MRBT maintains a state assignment $u: \mathcal{B}_{leaf} \rightarrow U$ initialized by $u_0$.
\end{definition}

% \begin{definition}\label{def:mrbt}
%     (masking reward behavior tree) Given a set of 
%     $\mathcal{L}$ logic formulas, discrete action
%     space $\mathcal{A}$, and BT $\mathcal{T}$, a 
%     masking reward behavior tree is a tuple 
%     $M\mathcal{R}_\mathcal{LAT} = \langle 
%     \mathcal{B}_{leaf}, \mathbf{U}, \mathbf{u}_0, \\
%     \delta_\mathbf{u}, \delta_\mathbf{r}, \delta_{\boldsymbol{\eta}} \rangle$, where 
%     $\mathcal{B}_{leaf}$ is a set of leaf MBRMs,
%     $\mathbf{U} = \prod_{b \in 
%     \mathcal{B}_{leaf}} U_b$ is the joint state 
%     of $\mathcal{B}_{leaf}$, $\mathbf{u}_0 \in \mathbf{U}$ is the joint initial state, 
%     $\delta_\mathbf{u}: \mathbf{U} \times 
%     2^\mathcal{L} \rightarrow \mathbf{U}$ is the
%     joint state-transition function,  
%     $\delta_\mathbf{r}: \mathbf{U} \times 
%     2^\mathcal{L} \rightarrow \mathbb{R}$ is the 
%     joint state-reward function, and 
%     $\delta_{\boldsymbol{\eta}}: \mathbf{U} 
%     \times 2^\mathcal{L} \rightarrow 
%     2^\mathcal{A}$ is a joint state-mask function.
% \end{definition}

An MRBT starts from the initial state assignment $u_0$, and at each timestep $t$, maintains a state assignment $u_t: \mathcal{B}_{leaf} \rightarrow U$. Given a label assignment $\sigma_{t+1} \subseteq \mathcal{L}$, a tick is propagated through $\mathcal{T}$ starting from the root, producing an ordered sequence of ticked leaves $\epsilon_t = (b_1, \dots, b_j)$, where $b_i \in \mathcal{B}_{leaf}$. The state assignment is updated by applying the transition function only to the ticked leaves:
\begin{equation}
    u_{t+1}(b) =
\begin{cases}
\delta_u^b\big(u_t(b), \sigma_{t+1}\big) & \text{if } b \in \varepsilon_t \\
u_t(b) & \text{otherwise}
\end{cases}
\end{equation}
The reward is the sum of rewards of the ticked leaves $r_{t+1} = \sum_{b \in \varepsilon_t} \delta_r^b\big(u_t(b), \sigma_{t+1}\big)$ and the action mask is the mask of the last ticked leaf $\eta_{t+1} = \eta^{b_j}$. 

% An MRBT $M\mathcal{R}_\mathcal{LAT}$ starts its MBRMs at joint state $\mathbf{u}_0$, and each subsequent step is in some joint state $\mathbf{u}_t \in \mathbf{U}$. At each step $t$, $M\mathcal{R}_\mathcal{LAT}$ receives as input a label assignment $\sigma_{t+1}$, which is a subset of logic formulas in $\mathcal{L}$ that are \emph{True}. The MRBT moves to the next joint state $\mathbf{u}_{t+1} = \delta_\mathbf{u}(\mathbf{u}_t, \sigma_{t+1})$ via the joint state-transition function. Internally, $\delta_{\mathbf{u}}$ propagates a \textit{tick} in $\mathcal{T}$ starting from the root. The MBRMs executed by the \textit{tick} are represented by a sequence of behaviors $\varepsilon_t = \{ b_1, b_2, ..., b_j \}$ where $b_i \in \mathcal{B}_{leaf}$ and $j$ is the number of ticked MBRMs. Therefore, the next joint state is equivalent to updating only the MBRMs ticked in $\varepsilon_t$ and can be written as: $\mathbf{u}_{t+1} = \delta_\mathbf{u}(\mathbf{u}_t, \sigma_{t+1}) = \langle \delta^b_u(u^b_t, \sigma_{t+1}) \text{ if } b \in \varepsilon_t \text{ else } u^b_t \mid b \in \mathcal{B}_{leaf} \rangle$. Similarly, the joint state-reward function aggregates the rewards from only the ticked MBRMs, written as: $\delta_\mathbf{r}(\mathbf{u}_t, \sigma_{t+1}) = \sum_{b \in \varepsilon_t} \delta^b_r(u^b_t, \sigma_{t+1})$. The joint state-mask function is equivalent to the action mask of the last ticked MBRM $b_j \in \varepsilon_t$, written as: $\delta_{\boldsymbol{\eta}}(\mathbf{u}_t, \sigma_{t+1}) = \eta^{b_j}$. The full algorithm is in Appendix~\ref{app: mrbt_update_step}.

\subsection{Masking Reward BT Template}
We introduce an MRBT template (Figure~\ref{fig:rbt_template}) to systematically construct an MRBT and a labeling function for a given environment $\mathcal{E}$ and task space $\mathcal{M}$. The template assumes tasks are decomposed into sequential subtasks involving navigation and object interactions, making it suitable for domains like robotic assembly and drone delivery.

The template inputs $k$ subtasks and $2k$ action masks and outputs an MRBT with labeling function $l(s,m)$. Each subtask $i$ has two masks, $\eta^{i,1}, \eta^{i,2} \in \mathcal{A}$. We first define logic formulas $\mathcal{L}$ and BT $\mathcal{T}$. $\mathcal{L}$ contains $2k$ formulas: $\psi^i$ for subtask  \textsc{completion} and $\phi^i$ for \textsc{object proximity} to the subtask object. The BT has a sequence root with $k$ subtree children, each comprising a fallback, a sequence, and three MBRMs (one condition and two actions) represented as $MB(\rho)$ that share a transition structure conditioned on $\rho \in \mathcal{L}$. The transition structure enables a positive reward to be outputted when $\rho$ is switched to \emph{True} and a reward penalty when switched to \emph{False}. Penalties signal to the agent that a subtask must be re-completed, and state changes in the MBRM trigger MRBT backtracking when a state moves from \emph{Success} to \emph{Failure} or \emph{Running}, ensuring the correct action mask is outputted.

The condition MBRM $MB(\psi^i)$ returns \textit{Success} if the subtask is complete and \textit{Failure} otherwise. The navigation MBRM $MB(\phi^i)$ returns \textit{Success} when the agent is near the object and \textit{Running} otherwise. The interaction MBRM $MB(\bot)$ always returns \textit{Running} until preempted by subtask completion or navigation. $MB(\psi^i)$ outputs the full action space, while $MB(\phi^i)$ and $MB(\bot)$ output $\eta^{i,1}$ (navigation) and $\eta^{i,2}$ (interaction), respectively. Finally, $\mathcal{L}$, $\mathcal{T}$, and $\mathcal{A}$ are used to derive an MRBT by Definition~\ref{def:mrbt}. 

Executing the MRBT requires evaluating the logic formulas in $\mathcal{L}$ to obtain a label assignment $\sigma_{t+1} = l(s_{t+1}, m)$ using a labeling function $l(s, m)$. The labeling function is expressed as the logic formulas in $\mathcal{L}$ satisfied from state $s \in \mathcal{S}$ and task $m \in \mathcal{M}$:

\begin{equation}\label{eq:labeling}
   l(s, m) =
    \big\{ \psi^i \mid 1 \le i \le k, \ (s, m) \models \psi^i \big\}
    \ \cup \
    \big\{ \phi^i \mid 1 \le i \le k, \ (s, m) \models \phi^i \big\}
\end{equation}

Given a task $m \in \mathcal{M}$ and current state assigment $u_t$, executing one tick of the MRBT with input $\sigma_{t+1}$ produces a sequence of ticked leaves $\epsilon_t$. The reward and action mask are then given by 
\begin{equation}\label{eq:rew_update}
r(\lambda_{t+1}) = \sum_{b \in \varepsilon_t} \delta_r^b(u_t(b), \sigma_{t+1})
\end{equation}
\begin{equation}\label{eq:mask_update}
    \mu(\lambda_{t+1}) = \eta^{b_j}
\end{equation}
where $b_j$ is the last leaf in $\epsilon_t$, and $\lambda_{t+1} = (\lambda_t, \sigma_{t+1})$.

% Given a task $m^i \in \mathcal{M}$, the reward function can be written using the joint state-reward 
% function, $r(\lambda_{t+1},  m^i) = \delta_\mathbf{r}
% (\mathbf{u}_t, \sigma_{t+1})$, where $\sigma_{t+1} = l(s_{t+1},  m^i)$ and
% $\mathbf{u}_t$ is the joint state derived from $\lambda_t$. Note that $\lambda_{t+1} = (\lambda_t, \sigma_{t+1})$, a concatenation.
% Similarly, the action mask function can be written using the joint state-mask function 
% $\mu(\lambda_{t+1}, m^i) = \delta_{\boldsymbol{\eta}}(\mathbf{u}_t, \sigma_{t+1})$.

% \begin{figure}[t]
%     \centering
%     \includegraphics[width=0.8\linewidth]{figs/approach/llm_assist_approach.pdf}
%     \caption{Automated pipeline to \textit{generate} and \textit{verify} MRBTs with an LLM and SMT solver.}
%     \label{fig:llm_assist_approach}
%     \vspace{-0.5cm}
% \end{figure}

\begin{exmp}\label{ex: lockedroom}
An MRBT for MiniGrid LockedRoom contains $k=4$ subtasks: \textit{Open key-room door, Pick up key, Unlock/open locked door, Reach goal}. For subtask $1$, $\psi^1$ checks if the door is open or occluded due to the agent being at the door position: 
$[door\_state_{room\_color} = \text{open } \lor door\_state_{room\_color} = -1]$ and $\phi^1$ checks if the Manhattan distance between the agent and door  is $\leq 1$ or the door is occluded: 
$manhattan\_distance(agent\_pos,  door\_pos_{room\_color}) \leq 1 \ \lor \ door\_state = -1$. The action mask $\eta^{1,1}$ contains actions \emph{\{left, right, forward\}} to navigate and $\eta^{1, 2}$ contains the actions \emph{\{left, right, toggle\}} to open the door.
\end{exmp}

\subsection{Verification of Masking Reward BTs}\label{sec: verification}
To verify the subtask logic formulas of an MRBT constructed from the template, we develop a symbolic model of the environment transition dynamics $\mathcal{E}^{sym}$ in an SMT-solver. We derive logical specifications derived from the formulas that are checked over task space $\mathcal{M}$ and trajectories $\tau = (s_0, \dots, s_{H-1})$ of horizon $H$ generated by the SMT-solver constrained by $\mathcal{E}^{sym}$. See Appendix~\ref{app: logic_specs} for a table of specifications.
% We compose specifications over the trajectory space $\tau \in \Gamma$ and task space $m \in \mathcal{M}$ that include logic formulas $g^m$, $\psi^i$, and $\phi^i$ (see Appendix~\ref{app: logic_specs} for specifications). We denote, $g^m_t$, $\psi^i_t$, and $\phi^i_t$ as timestep evaluations, and $\exists^{=N}$ as a counting quantifier checking existence of $N$ distinct solutions.
% Appendix~\ref{app: logic_specs} lists the specifications.

\noindent
\textbf{Completion:}
We verify that if the task is complete at final timestep $H-1$, then the subtask must have been completed. This ensures that $MB(\psi^i)$ will always transition to \emph{Success} at least once when the task is complete. Further, to prevent trivial solutions (i.e. $\psi^i = \top$) where the subtask will be complete in most trajectories and tasks, we verify the existence of $N$ distinct solutions where $\psi^i$ is never \emph{True}. 

\noindent
\textbf{Object Proximity:}
We verify that the agent is in proximity to the subtask object at the timestep immediately preceding subtask completion. This ensures that $MB(\phi^i)$ is at the \emph{Success} state at time $t$, before the subtask completes at $t+1$, and implies that $MB(\bot)$ is in the \emph{Running} state at $t$. Together, these conditions preserve the sequence of navigating to and interacting with the subtask object prior to completion in the MRBT template.
Further, we verify the existence of $N$ distinct solutions where $\phi^i$ is never \emph{True}.

\noindent
\textbf{Non-regressive Maximal Reward:}
To achieve maximal reward under the MRBT template, all subtasks must be completed. Assuming deterministic dynamics and non-regressive subtasks (i.e., $\psi^i$ can remain \emph{True}), we verify the existence of $N$ distinct solutions such that if the task is complete at final timestep $H-1$, then each $\psi^i$ persists to be \emph{True}. 

\subsection{Automating Generation, Verification, and Training}
We present an automated pipeline to generate and verify MRBTs and use them to train an agent policy over a task space. Given instructions, an MRBT template, and task-specific information, the pipeline uses an LLM to generate subtasks, action masks, and subtask logic formulas. The logic formulas are verified for satisfiability using an SMT-solver and symbolic environment model $\mathcal{E}^{sym}$ using the specifications (Section~\ref{sec: verification}). 

The LLM inputs a system prompt, including
instructions, the MRBT template, and task-specific 
information such as, $\mathcal{M}$, predicates from $\mathcal{S}$, and 
$\mathcal{A}$. The instructions prompt the 
LLM to generate subtasks and logic formulas for a BT. First, the LLM is prompted to generate a list of subtasks. 
For each subtask $i$, the LLM 
generates a completion logic formula $\psi^i$, an object 
proximity logic formula $\phi^i$, and action masks: $\eta^{i, 1}$, 
$\eta^{i, 2}$. The prompts enforce dependence on both state and task, ensuring robustness across the state and task spaces.
The resulting 
$\psi^i$, $\phi^i$, $\eta^{i, 1}$, $\eta^{i, 2}$, are used 
in the MRBT defined by the generated 
subtasks (illustration in Appendix~\ref{sec: nesy_rl_loop_example}). 

The SMT-solver inputs the generated logic formulas, checks for satisfiability of the logic specifications subject to $\mathcal{E}^{sym}$ for trajectories of size $H$, and either terminates with the specifications being satisfied or returns a debug prompt that is used to refine the generated logic formulas. The debug prompt includes the subtask logic formula that violated the specification and a failure description. For Non-Regressive Maximal Reward, it reports the first subtask violating persistence. For Completion and Object-Proximity correctness, an unsatisfiable specification yields a counterexample trajectory–task pair satisfying the negated specification, which is appended to the prompt.  

The MRBT is used as a reward function and action mask function in a neurosymbolic RL loop to learn $\pi^*$ using Equations~\ref{eq:rew_update} and \ref{eq:mask_update} (see example in Appendix~\ref{sec: nesy_rl_loop_example}).
% At step $t$ for task $m^i \in \mathcal{M}$, given next state $s_{t+1}$, label trace $\lambda_{t+1}$, an MRBT, and joint state $\mathbf{u}_t$ dependent on $\lambda_{t}$, the functions are:
% \begin{equation}
%     r(\lambda_{t+1},  m^i) = \delta_{\mathbf{r}}(\mathbf{u}_t, l(s_{t+1},  m^i))
% \end{equation}
% % and the action mask is:
% \begin{equation}
%     \mu(\lambda_{t+1}, m^i) = \delta_{\boldsymbol{\eta}}(\mathbf{u}_t, l(s_{t+1},  m^i))
% \end{equation}
% Both functions are used to optimize $\pi$ (see Appendix~\ref{sec: nesy_rl_loop_example} for integration example). 

%% file: chapters_v2/evaluation.tex
\section{Evaluation}

% overview

% \subsection{Environments}

% make sure to mention that these environments use discrete actions

We conduct a comprehensive evaluation that includes an analysis of the pipeline outputs and an ablation study that compares agent task success using four ablations of MRBT across five task spaces in two environments. We conduct ten experiments across all task spaces under both deterministic and stochastic dynamics. The deterministic setting evaluates whether MRBT reactivity improves training when subtask failures arise from the agent’s own actions. The stochastic setting evaluates whether MRBT reactivity benefits the agent under unpredictable changes (e.g., randomly dropping a key).  We use ChatGPT 5 as the LLM for generation of MRBTs and verify generated MRBTs using Z3 SMT-solver. 

\subsection{Environments}

The environments are \textit{MiniGrid}, with discrete states and actions, and \textit{MuJoCo Fetch} (\cite{gymnasium_robotics2023github}), with discrete actions and continuous states. \textit{MuJoCo Fetch} models simple assembly tasks in which a gripper agent must pick up a block and move it to a target position. For \textit{MuJoCo Fetch}, continuous position offsets were discretized and gripper actions reduced to open/close to allow discrete action masking. \textit{Stochastic} dynamics are modeled with a $0.05$ transition probability in both environments: in \textit{MiniGrid}, a picked-up key may drop to an adjacent cell; in \textit{MuJoCo Fetch}, the gripper may open unexpectedly, potentially dropping the block. Episodes may end upon task completion or at a timestep limit.

Task spaces for \textit{MiniGrid} include \textit{DoorKey}, \textit{LockedRoom}, and \textit{DroneSupplier}, while \textit{MuJoCo Fetch} includes \textit{PickAndPlace} and \textit{PickAndPlace2}. In \textit{DoorKey}, the agent picks up a key, opens a locked door, and reaches a goal; in \textit{LockedRoom}, it retrieves a key from one of six rooms to open a locked door to a goal; in \textit{DroneSupplier}, it collects a key from one of six boxes to unlock one of six doors. The grid in \textit{DroneSupplier} is extracted from a neighborhood map in AirSim. We relate keys to supplies and the locked door as a person in need. In \textit{PickAndPlace}, a gripper agent moves a block to a target, and in \textit{PickAndPlace2}, to one of two targets. More details on the environments and tasks are in Appendix~\ref{app: env_desc}.

\subsection{Automated Pipeline Results}

% describe input
% LLM, documents for the LLM (web scraping)

We use ChatGPT-5 for its strong reasoning and accessibility. Details on pipeline setup are in Appendix~\ref{app: pipeline_setup}.
% \vspace{0.1em}
% \noindent
% \textbf{Pipeline Results:}
% rbt structures
We report generated subtasks for each task space. Task space \textit{DoorKey} has subtasks \textit{Acquire Key, Open Door, Reach Goal}, task space \textit{DroneSupplier} has subtasks \textit{Open the box, Pick up the key, Open the door}, task space \textit{PickAndPlace} has  subtasks \textit{Pick Block, Move To Target}, task space \textit{PickAndPlace2} has subtasks \textit{Grasp Block, Move Block to Target}. 
% For task space \textit{LockedRoom}, subtasks are listed in Example~\ref{ex: lockedroom}. 
%For . For . For . 
% The generated MRBT for \textit{LockedRoom} is illustrated in Appendix~\ref{sec: nesy_rl_loop_example}.
% action masks
The action masks for each task space follow a similar pattern. The first action mask $\eta^{i, 1}$ for each subtask $i$ contains navigation actions to get near the subtask object. 
The second action mask $\eta^{i, 2}$ enables specific manipulation actions and can disable navigation. \textit{LockedRoom} subtasks and action masks are listed in Example~\ref{ex: lockedroom}.

% re-prompting example case
Z3 can be used to refine generated logic formulas. For $\psi^1$ in Example~\ref{ex: lockedroom}, the LLM initially generated the logic formula $[door\_state_{room\_color} = open]$. Z3 found the Non-regressive Maximal Reward specification to be unsatisfiable due to $\psi^1$ not persisting. After re-prompting the LLM with this failure, it correctly revised $\psi^1$ to include $door\_state_{room\_color} = -1$. In \textit{PickAndPlace2}, Z3 detected an object proximity counterexample where the block reached the target before the agent was evaluated in proximity using the object proximity logic formula; the LLM resolved this by increasing the distance thresholds. All correct specifications were verified in under ten minutes, with most taking less than one (see Appendix~\ref{app: verifyz-runtimes}).

\subsection{Ablation Study}\label{section:exp_setup}
% Algorithm comparison
% - Environment Reward
% - Procedure as Reward (Relate to how code as reward was given)
% - BT as Reward
% - BT as Reward + Action Mask
We train an agent policy $\pi(s, m)$ using neurosymbolic RL with four MRBT ablations, two of which are baselines, within each task space and evaluate performance based on the task \textit{success rate} of the agent. 
% Let $E_{[t: t + I]}$ be a sequence of episodes between step $t$ and $t+I$ during training, where each episode $i$ contains the task $m_i = \langle v_i, \ell_i, g_i \rangle$ and terminal state $s_i^f$. 
The \textit{success rate}
% \begin{equation}
%     \text{SR}_{[t: t+ I]} = \frac{\sum_{\big\langle m_i, s_i^f \big\rangle \in E_{[t: t + I]}} g_i\big(s_i^f\big)}{| E |}
% \end{equation}
% where $I = 2048$. Intuitively, $\text{SR}_{[t: t+ I]}$ 
is the proportion of tasks completed from step $t$ to $t + I$, where $I=2048$ timesteps. 
% Although faster agents may complete more tasks per interval, the success rate is still valid as it is normalized over number of episodes. 

The ablation methods of MRBT are as follows:

%\begin{itemize}
%\item 
\noindent
1) \textit{Task} - (Baseline) This baseline provides a binary reward if a task is complete.
%    \item 

\noindent
2) \textit{Procedure} - (Baseline) This baseline is based on the reward function in VLM-CaR, that is a procedure of reward code for evaluating completion of each subtask. Our implementation is a modified MRBT where  each subtask reward is issued only once (i.e. no reactive backtracking) and there is no action masking.
%\item 

\noindent
3) \textit{RBT} - (Ours) The reward function uses an MRBT, but no action masking.
%\item 

\noindent
4) \textit{MRBT} - (Ours) Reward function and an action mask function use MRBT.
%\end{itemize}

\textit{Procedure}, \textit{RBT}, and \textit{MRBT} use a shared MRBT structure and include the \textit{Task} reward as additional feedback. Training algorithm and hyperparameter details are in Appendix~\ref{app: training_setup}.

% We exclude evaluation of MBT (MRBT without rewards) and \textit{Procedure} with action masks. Generating an MRBT from an RBT requires additional effort to produce action masks, and the same inputs used to generate action masks (subtasks and code) for MBT also define the rewards for the RBT. Therfore, MBT and MRBT are functionally equivalent, making it more practical to use MRBT for fully enabled feedback. For \textit{Procedure} with action masks, the action masks may not align with the correct subtasks, providing no clear benefit. For example, in \textit{DoorKey}, due to a lack of reactivity, \textit{Procedure} might incorrectly output the action mask for opening the door after the key has dropped, when the correct action mask should be to pick up or navigate to the key.

\begin{figure}[t]
    \centering
    
    \begin{subfigure}
        \centering
        \includegraphics[width=0.23\linewidth]{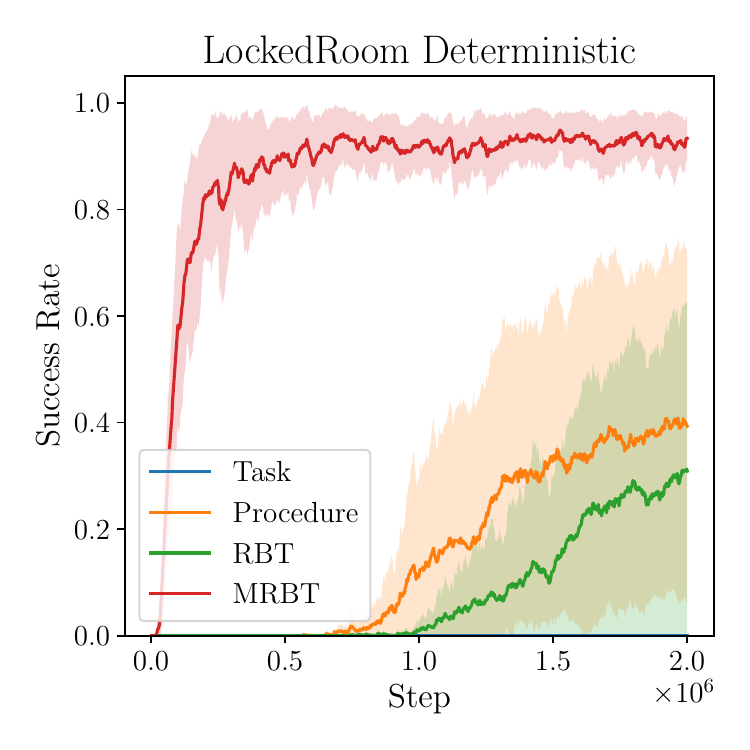}
    \end{subfigure} %
    \begin{subfigure}
        \centering
        \includegraphics[width=0.23\linewidth]{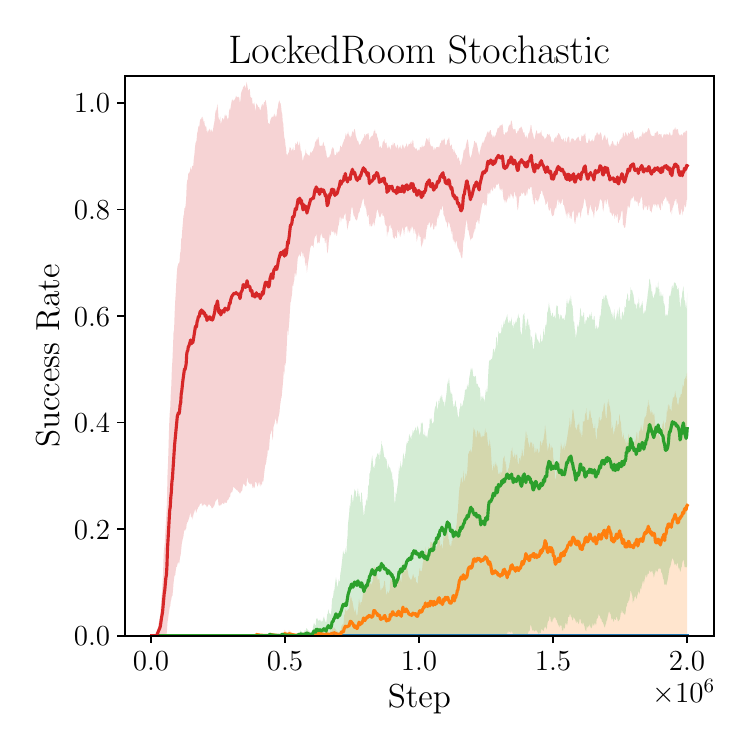}
    \end{subfigure} %
    \begin{subfigure}
        \centering
        \includegraphics[width=0.23\linewidth]{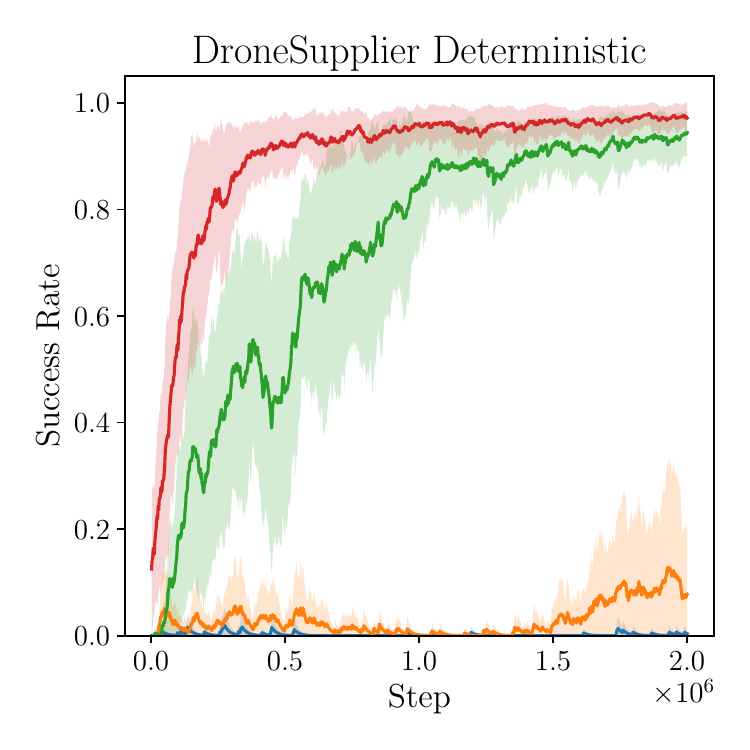}
    \end{subfigure} %
    \begin{subfigure}
        \centering
        \includegraphics[width=0.23\linewidth]{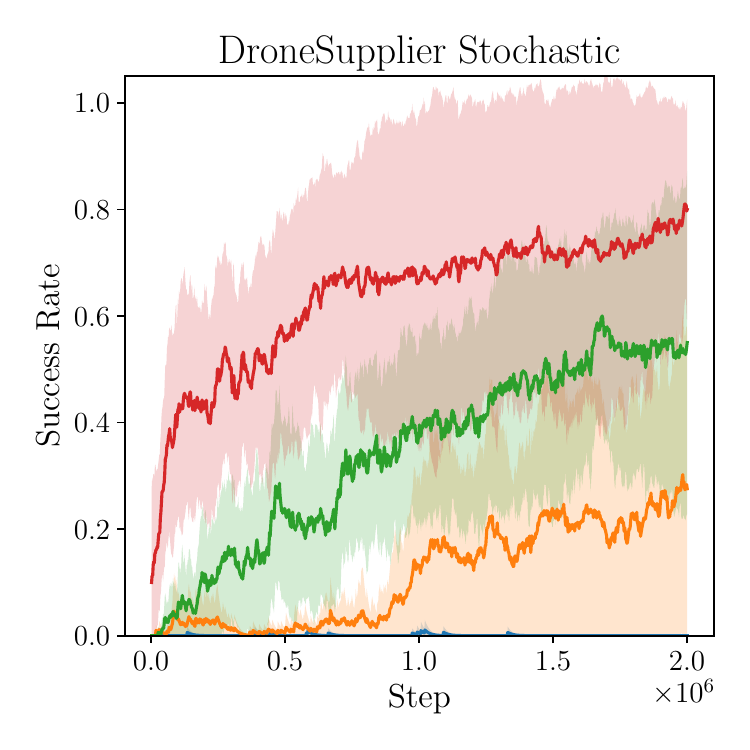}
    \end{subfigure} %
    \begin{subfigure}
        \centering
        \includegraphics[width=0.23\linewidth]{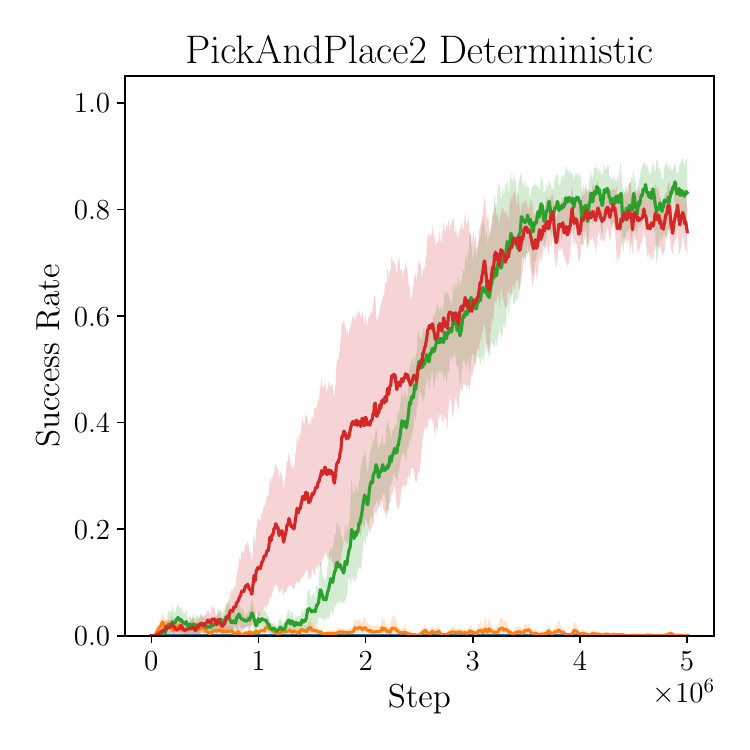}
    \end{subfigure} %
    \begin{subfigure}
        \centering
        \includegraphics[width=0.23\linewidth]{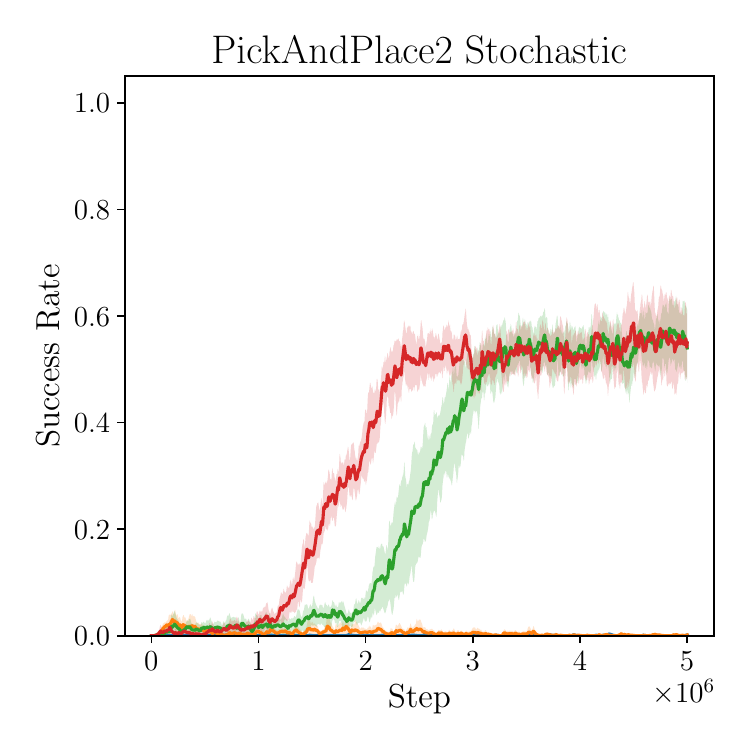}
        \phantomcaption
        \label{fig:pickandplace2_stoch}
    \end{subfigure}
    
    \caption{Success rate of agents during training for $4$ random seeds with \textit{deterministic} and \textit{stochastic} dynamics; Solid lines and shaded region represent mean and standard deviation.}
    \label{fig:minigrid}
    \vspace{-0.5cm}
\end{figure}

% doorkey
% lockedroom
% drone supplier
% pick and place
% pick and place 2
\vspace{0.1em}
\noindent
\textbf{Ablation Study Results:} 
\textit{MRBT} results in consistently high average task success across all experiments, whereas other ablations exhibit lower or more variable performance (Figure~\ref{fig:minigrid}). Using \textit{Task} results in near-zero task success across experiments, demonstrating the need for task decomposition. Specifically in \textit{LockedRoom}, the most complex task space with the largest number of subtasks ($k=4$) and task variants ($36$), the action masks significantly improve average task success ($ \geq 80\%$) over methods without action masking ($ \leq 70\%$) such as \textit{RBT}. In \textit{PickAndPlace2}, using \textit{Procedure} results in near-zero task success, whereas \textit{MRBT} and \textit{RBT} lead to $>60\%$ task success highlighting the need for reactive rewards to guide agent learning. Using action masks in \textit{MRBT} leads to improved learning efficiency and either comparable or higher average task success than using \textit{RBT}. A limitation of the action masks is that their bias can restrict the agent from learning a more optimal solution, as seen in \textit{PickAndPlace2} where there are instances the average task success rate of \textit{RBT} is higher than \textit{MRBT}. However, we find this degradation to be minimal, as the standard deviations of task success across four training runs overlap, and in all experiments \textit{MRBT} achieves consistently high task success. Results for \textit{DoorKey} and \textit{PickAndPlace} are in Appendix~\ref{app: more_exps}.

%% file: chapters_v2/advantages.tex
\section{Advantages of MRBTs}
%\vspace{0.1cm}
\noindent
\textbf{Transferability:} We evaluate the transferability of the \textit{DroneSupplier} agent 
policy in \textit{MiniGrid} to a realistic quadcopter simulator: 
AirSim. Since AirSim does not support 
\begin{wraptable}{r}{0.4\linewidth}
%\vspace{-1em}
\centering
\tiny
\begin{tabular}{|c|c|c|}
\hline
\textbf{Method} &
\textbf{\makecell{Success Rate \\ (Deterministic)}} &
\textbf{\makecell{Success Rate \\ (Stochastic)}} \\ \hline
Task & $0.00$ & $0.00$ \\ \hline
Procedure & $0.05$ & $0.57$ \\ \hline
RBT & $0.94$ & $0.45$ \\ \hline
MRBT & $0.97$ & $0.94$ \\ \hline
\end{tabular}
\caption{Task success rate for agent policies transferred to AirSim.}
\label{tab:airsim_transfer}
%\vspace{-0.6cm}
\end{wraptable}
 manipulator actions, we focus on evaluating navigation and represent interaction actions as transitions in \textit{MiniGrid}. 
Using 
AirSim’s API, we implement forward motion via \texttt{moveToPosition}
and directional changes via \texttt{moveByVelocity}. We compare the task success rates of the ablation methods, 
in Section~\ref{section:exp_setup}, over $100$ episodes. Results in  
% We evaluate under \textit{deterministic} and \textit{stochastic} dynamics. 
%The results in Table~\ref{tab:airsim_transfer} show that \textit{MRBT} still leads to higher task success rate than other ablation methods. 
Table~\ref{tab:airsim_transfer} show that the task success rates under MRBT outperforms all ablation methods.
% An advantage of MRBTs over ablations is their ability to emit action masks to the agent policy during deployment. Consequently, a policy trained with an MRBT in \textit{MiniGrid} continues to attain high task success rate when transferred to  Microsoft AirSim.

\noindent
\textbf{Modularity:}
The MRBT integrates action masking and is more modular than a hierarchical reward machine (HRM) (\cite{furelos2023hrms}) with comparable reward logic (Appendix~\ref{app: hrm}). For k=3 subtasks, the MRBT comprises 16 behaviors, 18 RM states, and 36 RM edges, while the HRM has 13 states and 24 edges. Although the HRM is structurally more compact, the MRBT is more modular: adding a subtask to the MRBT requires $\mathcal{O}(1)$ additional storage by attaching a new subtree to the root, whereas adding a subtask to the HRM requires $\mathcal{O}(k)$ storage for back edges to prior states. Prior work similarly shows that BTs scale more modularly than finite-state machines as task complexity increases (\cite{iovino2023programmingeffort, iovino2025comparingbtsandfsms, olsson2016behavior}).

% compare against HRM
% The MRBT integrates action masking and is more modular compared to a hierarchical reward machine (HRM)~\cite{furelos2023hrms} with similar reward logic (figure in Appendix~\ref{app: hrm}). An HRM is a reward machine with transitions that invoke subtask reward machines.
% % nodes, edges
% Given $k=3$ subtasks, the MRBT contains $16$ behaviors, $18$ RM states, and $36$ RM edges, whereas the HRM contains $13$ RM states and $24$ RM edges. 
% % modularity: insertion, deletion
% Although the HRM is structurally more compact than the MRBT, the MRBT is more modular. Inserting an additional subtask into the MRBT takes $O(1)$ additional storage to attach a new subtask subtree to the root. However, inserting a subtask into the HRM takes $O(k)$ additional storage to insert back edges to prior RM states.  
% Previous studies also show that BTs are more modular compared to finite-state machines as task complexity increases~\cite{iovino2023programmingeffort, iovino2025comparingbtsandfsms, olsson2016behavior}.

\noindent
\textbf{Verifiability:} We use Z3 to develop $\mathcal{E}^{sym}$ and verify logic specifications to provide theoretical guarantees for the MRBT. For \textit{MiniGrid}, we develop $\mathcal{E}^{sym}$ to match the true dynamics in $\mathcal{E}$, allowing us to conclude that the verified specifications hold during training for trajectories of size $H$. For \textit{MuJoCo Fetch}, $\mathcal{E}^{sym}$ abstracts $\mathcal{E}$. Although $\mathcal{E}$ is not fully modeled in $\mathcal{E}^{sym}$, we see from the results that the generated MRBTs improve training efficiency. In cases where $\mathcal{E}^{sym}$ is not available, expert demonstrations can be used to test logic formulas and can be useful priors for action masks (see Appendix~\ref{app: demonstrations}).

%% file: chapters_v2/conclusion.tex
\section{Conclusion}
We introduce masking reward behavior trees (MRBTs), used as reward and action mask functions for compositional tasks. We design an MRBT template and logical specifications to construct and verify MRBTs for sequential object-interaction subtasks. We design a pipeline to automatically generate and verify MRBTs using an LLM and SMT-solver and use the MRBTs in a neurosymbolic RL training loop. We generate five MRBTs using the pipeline and demonstrate how the SMT-solver was used to refine the generated output. An ablation study shows that MRBTs improve learning efficiency and task success over ablations, with  the largest gain in task success in the most complex task space. We further show that MRBTs offer transferability, modularity, and verifiability. Future work will explore generalizing the MRBT template. Generative AI was used to improve writing quality (ChatGPT-5) and to assist with code development (Claude Sonnet 4.5 (\cite{anthropic2025claude45})). This material is based upon work sponsored by the Air Force Research Lab (AFRL) and DARPA. Any opinions, findings, and conclusions or recommendations expressed in this material are those of the authors and do not necessarily reflect the views of AFRL or DARPA.

%% file: chapters_v2/background_related.tex
\section{Related Work}\label{app: related_work}
% LLMs for reward shaping
Recent methods leverage LLMs and VLMs to automate reward and action mask design. VLMs such as OpenAI CLIP (\cite{radford2021learningtransferablevisualmodels}) is used to compare embeddings of goal descriptions and current states, with high cosine similarity indicating goal completion. MineDojo (\cite{fan2022minedojo}) fine-tunes CLIP for task success in \textit{Minecraft} (\cite{kanervisto2022minerldiamond2021competition}), while trajectory embeddings can also be used for evaluation (\cite{sontakke2023roboclip}). Flamingo (\cite{alayrac2022flamingo}) has been fine-tuned for visual question answering to detect mission completion (\cite{du2023flamingosuccessdetector}), though fine-tuning and repeated inference during RL training incurs computational cost. Other approaches, including ours, generate code or symbolic rewards, requiring a pre-trained LLM or VLM during design or between training runs. Eureka (\cite{ma2023eureka}) uses an LLM to code and edit reward functions based on agent performance, while VLM-CaR (\cite{venuto2024codeasreward}) generates procedural subtask reward functions from initial and final image frames of a complex mission and validates them via a testing process. VLM-CaR effectively generates and tests reward code for complex missions but lacks support for varying missions in a mission space and reactivity to handle stochastic dynamics. LLMs can also produce multi-faceted rewards for richer feedback (\cite{Qu2025latentrewardllm}). Further, policies learned from LLM generated rewards were shown to support sim-to-real transfer for robotic tasks (\cite{qi2025roboticskillrewardsllms}). Symbolic rewards such as reward machines and logic specifications have likewise been augmented by LLMs: reward machines can be generated or optimized via few-shot LLM prompts encoding domain knowledge (\cite{alsadat2025llmsrm, alsadat2025llmswarmrm}), and LLM generated explanations can produce linear temporal logic formulas from expert demonstrations to construct reward machines or direct reward functions (\cite{gupta2024integratingexplanationslearningltl, camacho2019ltlrewards, tuli2022ltlinstructionrewards}). Less work has focused on LLMs for action masking with one work using LLMs to generate suboptimal policies as guidance to dynamically restrict a continuous action space (\cite{zhao2025camelcontinuousactionmasking}). Our proposed approach further expands the literature on methods for using LLMs to design action masks.

% why is ours using documents instead of images
% primarily because it requires less data to represent the mission space as text than images.
The prior work uses LLMs for single tasks to generate rewards and action masks, whereas our work focuses on designing joint reward and action mask functions that are reactive to subtask failure and generalize to variants of a task. Additionally, we verify specifications derived from the subtask logic formulas using an SMT solver, providing formal guarantees that are not attainable through testing with expert demonstrations.

%% file: appendix/nesy_rl_loop.tex
\section{Automated Pipeline More Details}\label{sec: nesy_rl_loop_example}
\begin{figure}[H]
    \centering
    \includegraphics[width=0.8\linewidth]{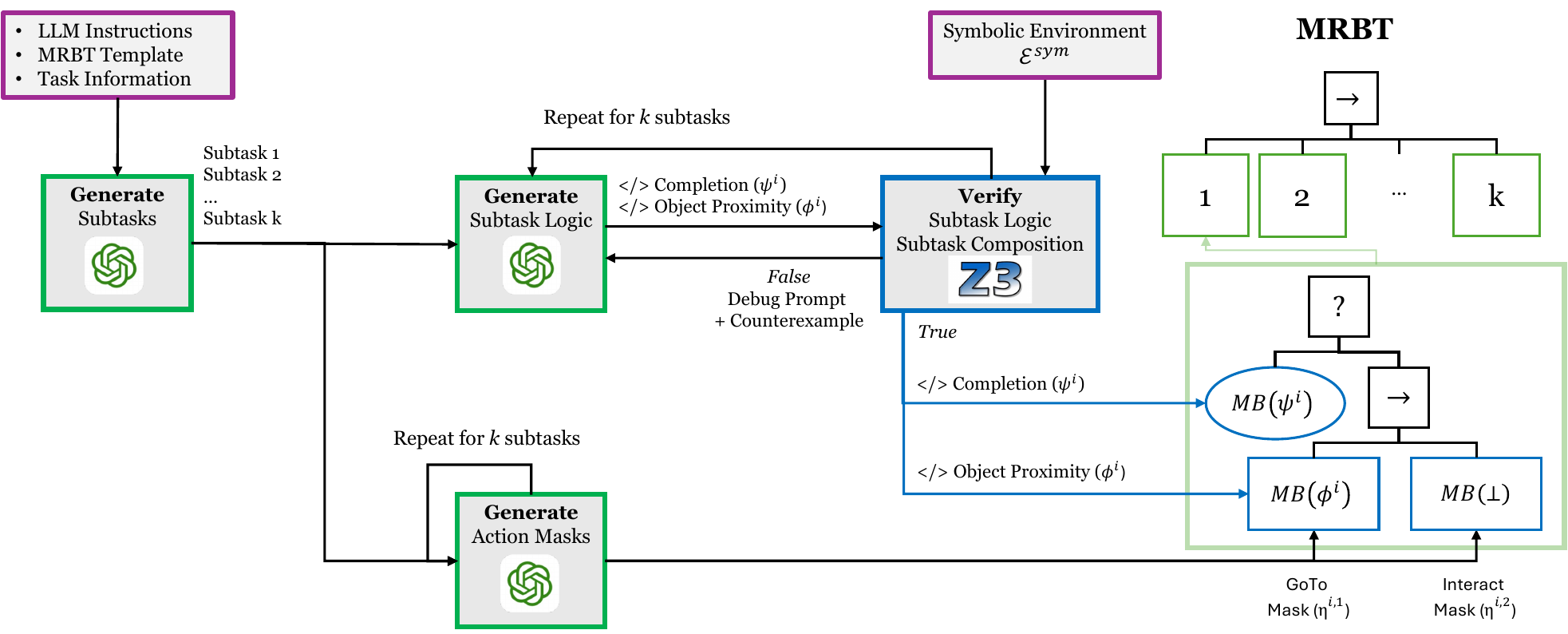}
    \caption{Automated pipeline to \textit{generate} and \textit{verify} MRBTs with an LLM and SMT solver.}
    \label{fig:llm_assist_approach}
    \vspace{-0.5cm}
\end{figure}

\begin{figure}[H]
    \centering
    \includegraphics[width=0.5\linewidth]{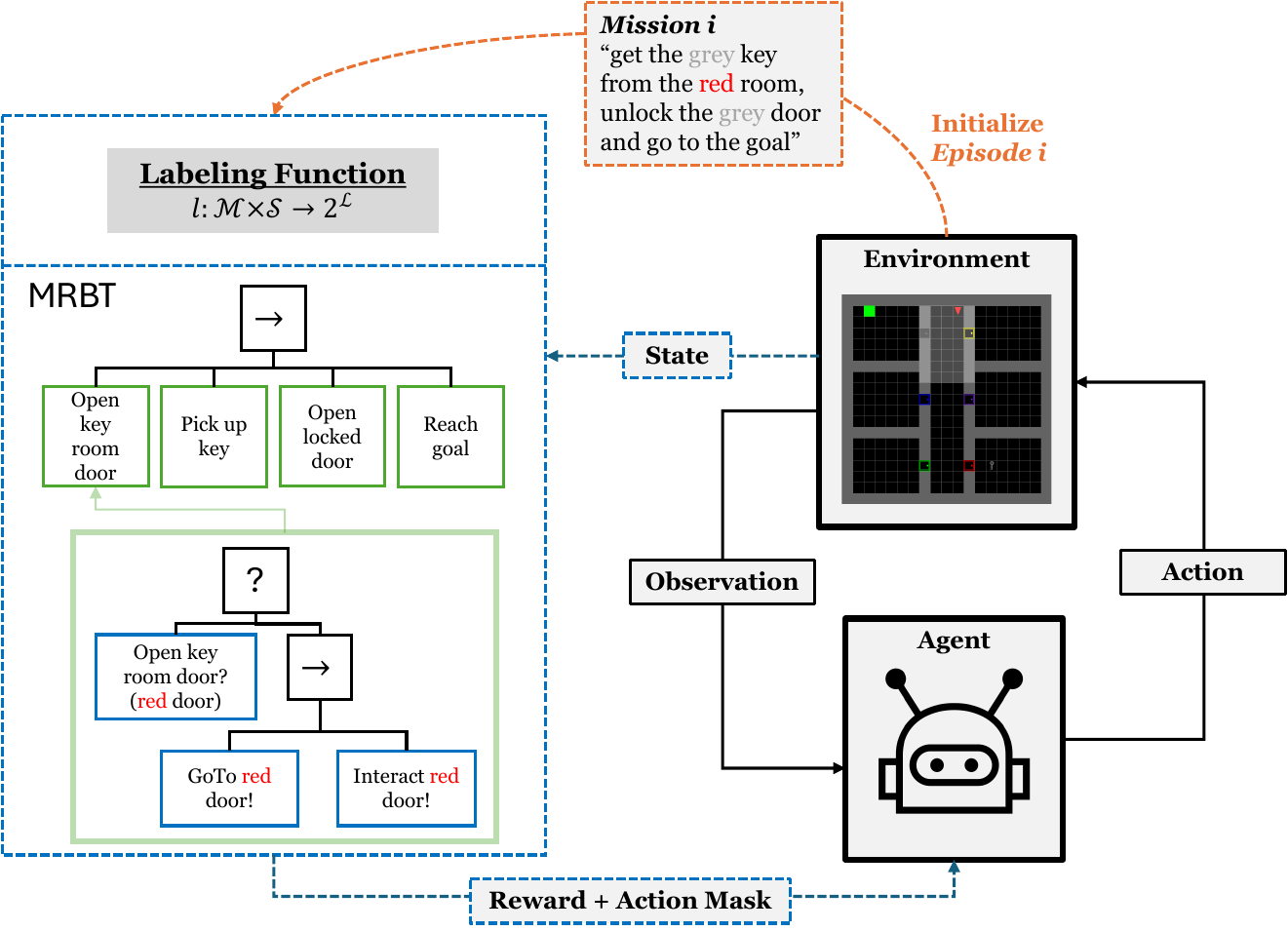}
    \caption{Integration of MRBTs into a neurosymbolic RL loop.}
    \label{fig:nesyrlloop}
\end{figure}

An example of the neurosymbolic RL loop for a single mission of \textit{MiniGrid LockedRoom} is in Figure~\ref{fig:nesyrlloop}. 
In this example, the color of the door (``red'') in subtask 1, \textit{Open key room door}, is determined by the episode mission, which specifies \textit{``...the red room...''}.

%% file: appendix/logic_specifications.tex
\section{Logic Specifications for Verification}\label{app: logic_specs}
We compose specifications over the trajectory space $\tau \in \Gamma$ and task space $m \in \mathcal{M}$ that include logic formulas $g^m$, $\psi^i$, and $\phi^i$. We denote, $g^m_t$, $\psi^i_t$, and $\phi^i_t$ as timestep evaluations, and $\exists^{=N}$ as a counting quantifier checking existence of $N$ distinct solutions.

\begin{table}[H]
\centering
\tiny
\begin{tabular}{m{0.4\linewidth} | m{0.6\linewidth}}

\textbf{Specification} & \textbf{Logic Formula} \\
\hline
% \midrule
Completion Correctness &
\[
\forall (\tau \in \Gamma, m \in \mathcal{M}^{sym}) \;
g^m_{H-1} \implies \bigvee_{t=0}^{H-1} \psi^i_t
\]
\\
\hline
Completion Non-Triviality &
\[
\exists^{=N} (\tau \in \Gamma, m \in \mathcal{M}) \;
\bigwedge_{t=0}^{H-1} \sim \psi^i_t
\]
\\
\hline
Object Proximity Correctness &
\[
\forall (\tau \in \Gamma, m \in \mathcal{M}) \;
\bigwedge_{t=0}^{H-2}
(\sim \psi^i_t \wedge \psi^i_{t+1})
\implies \phi^i_t
\]
\\
\hline
Object Proximity Non-Triviality &
\[
\exists^{=N} (\tau \in \Gamma, m \in \mathcal{M}) \;
\bigwedge_{t=0}^{H-1} \sim \phi^i_t
\]
\\
\hline
Composition Persistence &
\[
\exists^{=N} (\tau \in \Gamma, m \in \mathcal{M}) \;
g_{H-1} \implies
\bigwedge_{t=0}^{H-2}
\bigwedge_{i=1}^{k}
(\psi^i_t \implies \psi^i_{t+1})
\]
\\
\hline
\end{tabular}
\caption{Logic specifications used to verify subtasks.}
\label{tab:logic-specifications}
\end{table}

%% file: appendix/env_task_descriptions.tex
\section{Environment and Task Descriptions}\label{app: env_desc}

\subsection{MiniGrid}\label{section:minigrid}

\textit{MiniGrid} (\cite{MinigridMiniworld23}) is a discrete state, discrete action environment where an agent (red triangle) can pick up and interact with objects. The discrete state space $\mathcal{S}$ is a $N \times N$ grid where each grid cell is a tuple $(OBJECT ID, \allowbreak COLOR, STATE)$. $STATE$ stores data for the direction of the agent or if an object is open, closed, or locked. The observation space $\mathcal{O}$ is a $J \times J, J \leq N$ field of view of the state. Predicates of $\mathcal{S}$ contain the positions of objects (e.g. keys, doors, agent, goal, ...) and agent direction. The action space is \textit{\{left, \allowbreak right, \allowbreak forward, pickup, drop, toggle, done\}}. Stochastic dynamics are modeled as a $0.05$ probability in the transitions probability function $p$ that the key will drop to an adjacent cell after picked up. The color values are \textit{"red", "green", "blue", "purple", "yellow", "grey"}.

\textit{DoorKey.}
% discrete state, action
The agent must pick up a key, open a locked door, and then travel to the goal on the other side of a wall. The grid we consider is size $16 \times 16$ that has one wall and door that separates two rooms, where the left room has a key, and the right room has a goal. The task space is $\mathcal{M}_{dk} = \langle \ell_{dk}, \mathcal{V}_{dk}, \mathcal{G}_{dk} \rangle$.
$\ell_{dk} =$ textit{“use the key to open the door and then get to the goal”}, $\mathcal{V}_{dk} = \emptyset$, and $\mathcal{G}_{lr}$ contains one function that checks if the agent is at the same location as the goal. The maximum number of steps is $500$.

\textit{LockedRoom.}
% discrete state, action
The agent must pick up a key in one of six room, then open a locked door for another room to reach a goal. The grid is $19 \times 19$ with walls and doors to separate rooms and one key, one locked door, and one goal. The task space is $\mathcal{M}_{lr} = \langle \ell_{lr}, \mathcal{V}_{lr}, \mathcal{G}_{lr} \rangle$. $\ell_{lr} =$ \textit{“get the \{lockedroom colour\} key from the \{keyroom colour\} room, unlock the \{door colour\} door and go to the goal”}, $\mathcal{V}_{lr} = \{ \textit{lockedroom colour}, \textit{keyroom colour}, \allowbreak \textit{door colour} \}$, and $\mathcal{G}_{lr}$ contains one logic formula that checks if the agent is at the same location as the goal. The maximum number of steps is $190$.

\textit{DroneSupplier.}
% discrete state, action
The agent must pick up a key from one of six boxes and use the key to unlock and open one of six doors. The grid is a $25 \times 25$ slice of a neighborhood in Microsoft AirSim (\cite{airsim2017fsr}) where each object above an altitude of $5$ meters in the neighborhood is represented as a wall. The key is equivalent to supplies and the locked door is equivalent to a car or person in need. 
% This allows for an agent trained in DroneSupplier to be deployed to Microsoft AirSim on a realistic quadcopter. 
The task space is $\mathcal{M}_{ds} = \langle \ell_{ds} \mathcal{V}_{ds}, \mathcal{G}_{ds} \rangle$. $\ell_{ds} =$ \textit{“open the \{ box color \} box, pick up the key, then open the \{ door color \} door”}, $\mathcal{V}_{ds} = \{ \textit{box color}, \allowbreak \textit{door color} \}$, and $\mathcal{G}_{ds}$ contains logic formulas that check if the \textit{door color} door is open. The maximum number of steps is $500$.

\subsection{MuJoCo Fetch}

\textit{MuJoCo Fetch} is a \textit{Gymnasium-Robotics} (\cite{gymnasium_robotics2023github}) environment we modified as a continuous state, discrete action environment where the agent (gripper) must pick up a block and move it to targets. The continuous state space, $\mathcal{S}$, contains a $21-dimensional$ vector for gripper and block dynamics, a $3h-dimensional$ desired target vector for $h$ possible target positions. Predicates of $\mathcal{S}$ contain the gripper and block positions, linear velocity, target positions, and gripper finger displacement. The observation space is the state space, $\mathcal{O} = \mathcal{S}$. The discrete action space is \textit{\{left, \allowbreak right, \allowbreak forward, backward, up, down, open gripper, close gripper\}}. Stochastic dynamics are modeled as a $0.05$ probability in the transition probability function $p$ that the gripper will open when closed, potentially leading to the block dropping. The objects in $\mathcal{X}$ are \textit{"block", "target", "agent"}. The \textit{color} values for targets are \textit{"green", "yellow"}. For both task spaces, the maximum number of steps is $100$.

\textit{PickAndPlace.}
% continuous state, discrete action
The agent must pick up a block from a table and move it to a target position above or on the table. The task space is $\mathcal{M}_{pp} = \langle \ell_{pp}, \mathcal{V}_{pp}, \allowbreak \mathcal{G}_{pp} \rangle$. $\ell_{pp} =$ \textit{“pick up and move the block to the target location”}, $\mathcal{V}_{pp} = \emptyset$, and $\mathcal{G}_{pp}$ contains one logic formula that checks if the block is within $5$cm of the target. 

\textit{PickAndPlace2.}
% continuous state, discrete action
Same as \textit{PickAndPlace}, but with two target positions to select from. The task space is $\mathcal{M}_{pp2} = \langle \ell_{pp2}, \mathcal{V}_{pp2}, \allowbreak \mathcal{G}_{pp2} \rangle$. $\ell_{pp2} =$ \textit{“pick up and move the block to the \{ target color \} target location”}, $\mathcal{V}_{pp2} = \{ \textit{target color} \}$, and $\mathcal{G}_{pp}$ contains two logic formulas that check if the block is within $5$cm of the \textit{green} or \textit{yellow} target.

%% file: appendix/experiment_setup.tex
\section{Experimental Setup}

\subsection{Automated Pipeline Details}\label{app: pipeline_setup}
The pipeline interfaces with ChatGPT-5 using the Vanderbilt Amplify Platform (\cite{gaiin_platform_github}), an open-source web platform for hosting LLM services. However, any API could be adapted into the pipeline with a similar expected output. System prompts include manually curated LLM instructions and the MRBT template. For \textit{MiniGrid}, the task space $\mathcal{M}$ and action space $\mathcal{A}$ were web-scraped from the official documentation, while symbolic variables used for predicates from state space $S$ were Python Z3 variable definitions. In our custom \textit{MiniGrid} task space, \textit{DroneSupplier}, $\mathcal{M}$ was manually specified, with the task renamed to \textit{Unlock} to reflect its intent. For \textit{MuJoCo Fetch},  $\mathcal{M}$, predicates of $\mathcal{S}$, and $\mathcal{A}$ were manually specified similar to \textit{MiniGrid}. 

The prompts used for generating subtask logic formulas and action masks were similar for both environments. For generating subtask logic formulas, the LLM was instructed to generate a Python function compatible with inputting a vector of Z3 state variables and mission variables and outputting a Z3 expression. These functions could then be integrated into Z3 constraints for verification with symbolic environment $\mathcal{E}^{sym}$. Other details were included in prompting the model to exclude Manhattan or Euclidean distances from generated formulas, which can significantly increase verification time in continuous-state environments like $MuJoCo Fetch$. To generate action masks, the LLM was instructed to fill in vectors of size $|\mathcal{A}|$ with the available actions for the $2k$ action masks. 

The time horizon $H=25$ timesteps for verification in both environments. For finding $N$ distinct trajectories, we add constraints to prevent the initial state from being equal to previous initial states. For computational efficiency, we allow for setting a wallclock timeout $T_{timeout}$. The verifier attempts to find a counterexample, if none are found within $T_{timeout}$ the result is inconclusive. In such cases, the pipeline assumes correctness, interpreting a timeout without counterexamples as evidence that violations are difficult to find; thus, the logic formulas are expected to provide useful feedback for guiding training despite potential edge-case failures. $T_{timeout}=900$ seconds (15 minutes) for our experiments. While no verification step reached this timeout (see Table~\ref{tab: verify_times}), it serves as a practical safeguard when verifying logic formulas in more complex environments with potentially intractable verification times.

\subsection{Training Setup}\label{app: training_setup}
The RL algorithm for training is proximal policy optimization (PPO) with a learning rate of $3e-4$, $\gamma=0.99$, and the number of environment steps per optimization step is $2048$. For \textit{MiniGrid}, the agent policy consists of a feature encoder and a linear control head. The feature encoder comprises three convolutional layers with $16$, $32$, and $64$ filters of size $2 \times 2$, each followed by a ReLU activation. The output is flattened and passed through two additional fully connected layers of size $64$, that output action logits. For \textit{MuJoCo Fetch}, the agent policy consists of the linear control head with two fully connected layers of size $128$. 

%% file: appendix/verification_times.tex
\section{Verification Runtimes}\label{app: verifyz-runtimes}
We report the wallclock time to verify each specification in Section~\ref{sec: verification} using Z3 for each correct subtask logic formula generated by ChatGPT-5. The results are in Table~\ref{tab: verify_times}. The task spaces with the most time to verify are from \textit{LockedRoom} and \textit{DroneSupplier} because they have more task variants. All verification times are below the timeout $T_{timeout}=900$.

\begin{table}[H]

\centering
\caption{Verification time for logical specifications across five task spaces. Task spaces are \textit{DoorKey} (DK), \textit{LockedRoom} (LR), \textit{DroneSupplier} (DS), \textit{PickAndPlace} (PP), and \textit{PickAndPlace2} (PP2).}
\label{tab: verify_times}
\begin{tabular}{|c|c|c|c|c|c|}
\hline
\multirow{2}{*}{Specification} & \multirow{2}{*}{Env} & \multicolumn{4}{c|}{Subtask Verification Time (s)} \\ \cline{3-6} 
                        &                      & 1      & 2      & 3      & 4      \\ \hline
\multirow{5}{*}{Completion Correctness} 
                        & DK   & 3.18   & 1.72   & 0.93   & -    \\ 
                        & LR   & 199.21 & 51.03  & 14.53  & 1.04   \\ 
                        & DS   & 69.54  & 475.87 & 17.71  & -    \\ 
                        & PP   & 1.33   & 5.22   & -    & -    \\ 
                        & PP2  & 0.36   & 60.13  & -    & -    \\ \hline
\multirow{5}{*}{Completion Non-Triviality} 
                        & DK   & 17.74 & 25.21 & 24.11 & -   \\ 
                        & LR   & 61.74 & 61.17 & 71.42 & 43.21 \\ 
                        & DS   & 56.18 & 68.98 & 75.82 & -   \\ 
                        & PP   & 4.41  & 2.61  & -   & -   \\ 
                        & PP2  & 2.12  & 2.29  & -   & -   \\ \hline
\multirow{5}{*}{Object Proximity Correctness} 
                        & DK   & 3.54   & 8.96   & 3.08   & -   \\ 
                        & LR   & 243.77 & 164.40 & 132.32 & 49.41 \\ 
                        & DS   & 69.32  & 235.19 & 78.66  & -   \\ 
                        & PP   & 2.66   & 5.91   & -    & -   \\ 
                        & PP2  & 5.17   & 8.09   & -    & -   \\ \hline
\multirow{5}{*}{Object Proximity Non-Triviality} 
                        & DK   & 22.43 & 20.95 & 21.77 & -   \\ 
                        & LR   & 42.02 & 38.67 & 38.49 & 48.52 \\ 
                        & DS   & 55.59 & 78.17 & 60.54 & -   \\ 
                        & PP   & 3.39  & 2.37  & -   & -   \\ 
                        & PP2  & 2.40  & 1.94  & -   & -   \\ \hline
\multirow{5}{*}{Non-regressive Maximal Reward} 
                        & DK   & \multicolumn{4}{c|}{32.03}  \\ 
                        & LR   & \multicolumn{4}{c|}{437.73} \\ 
                        & DS   & \multicolumn{4}{c|}{359.79} \\ 
                        & PP   & \multicolumn{4}{c|}{11.06}  \\ 
                        & PP2  & \multicolumn{4}{c|}{8.23}   \\ \hline
\end{tabular}
\end{table}

%% file: appendix/exp_results.tex
\section{More Experiments Results}\label{app: more_exps}
\begin{figure}[H]
    \centering
    % First subfigure
    \begin{subfigure}
        \centering
        \includegraphics[width=0.25\linewidth]{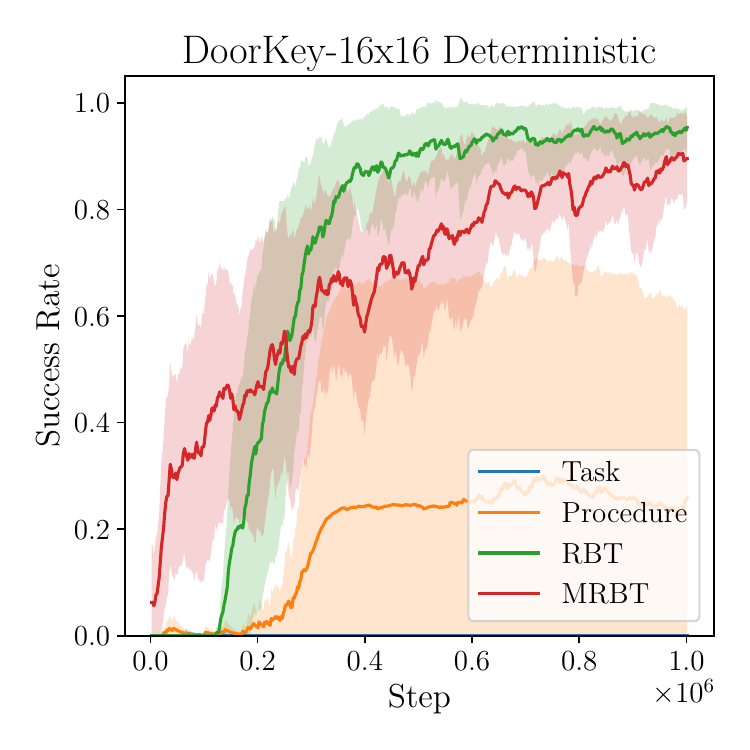}
    \end{subfigure} %
    % Second subfigure
    \begin{subfigure}
        \centering
        \includegraphics[width=0.25\linewidth]{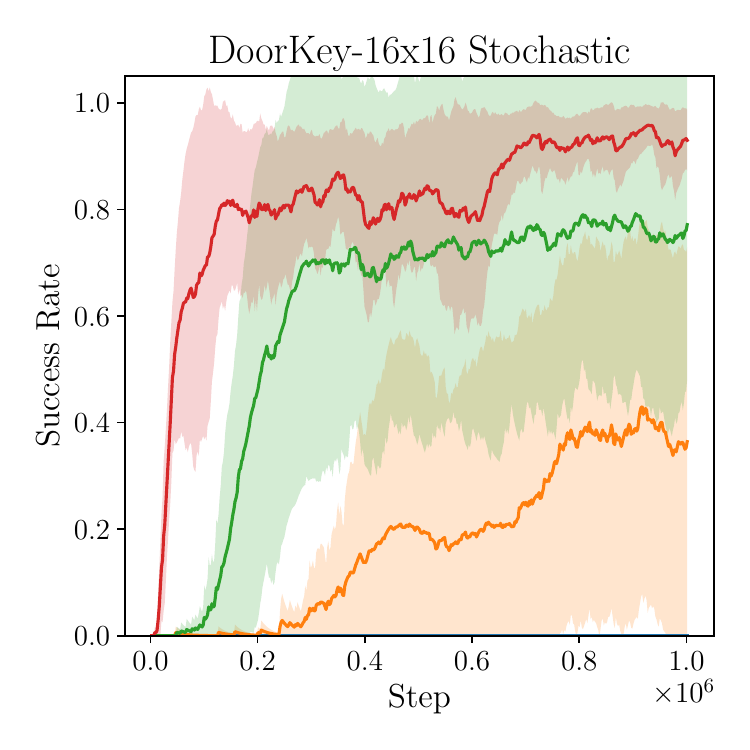}
    \end{subfigure} %
    % First subfigure
    \begin{subfigure}
        \centering
        \includegraphics[width=0.25\linewidth]{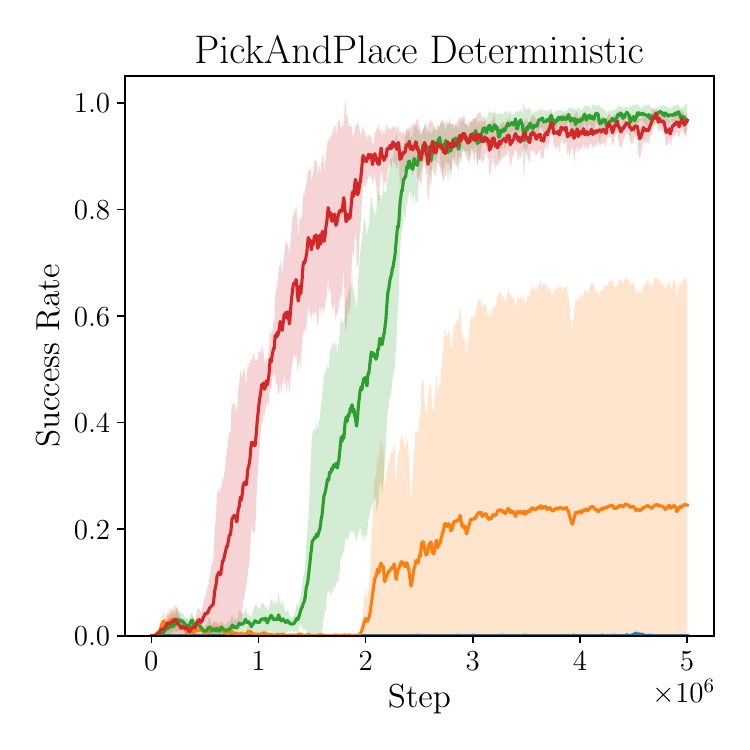}
    \end{subfigure} %
    % Second subfigure
    \begin{subfigure}
        \centering
        \includegraphics[width=0.25\linewidth]{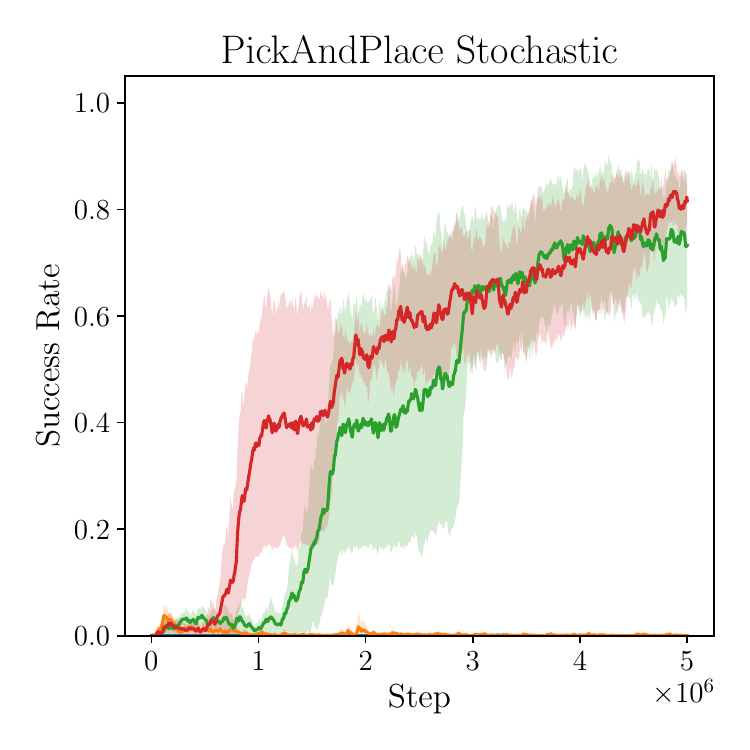}
    \end{subfigure}
    
    \caption{Task success rate of the agent during training for \textit{MiniGrid DoorKey} and \textit{MuJoCo Fetch PickAndPlace}.}
    \label{fig:more_exps}
\end{figure}

% In Figure~\ref{fig:more_exps}, the results from \textit{DoorKey} are similar to Section~\ref{section:exp_setup}. In \textit{PickAndPlace}, 

%% file: appendix/hrm.tex
\section{Hierarchical Reward Machine}\label{app: hrm}

\begin{figure}[h]
    \centering
    \includegraphics[width=0.8\linewidth]{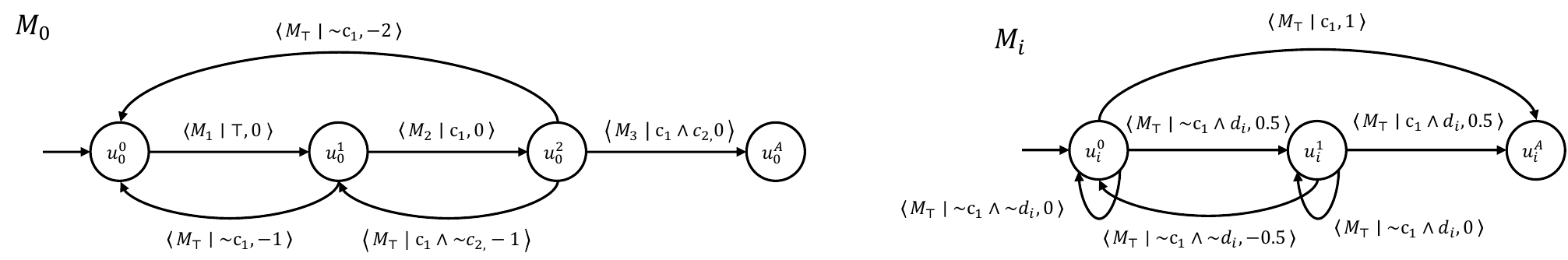}
    \caption{Hierarchical reward machine with three subtasks; the root RM is $M_0$, $u^A$ is an accepting state, $M_i$ are subtask RMs, and $M_\top$ is an RM that is $u^A$. Edges are labeled as $\langle M_i \mid e, r \rangle$ where $M_i$ is a RM called when exit condition $e$ is true, executing until it transitions to $u^A$, then outputting scalar reward $r$.}
    \label{fig:hrm}
\end{figure}

%% file: appendix/using_demonstrations.tex
\section{Testing MRBTs using Demonstrations}\label{app: demonstrations}
The specifications from Section~\ref{sec: verification} can be tested using expert and random demonstrations. The demonstrations are trajectories of states of arbitrary length constrained by $\mathcal{E}^{sym}$ and a task. The demonstrations are collected by an expert policy designed for $\mathcal{E}$ with deterministic dynamics. The Completion Correctness, Object Proximity Correctness, and Non-regressive Maximal Reward specifications are checked to see if they are satisfied for all trajectories or $N=10$ demonstrations. For Completion Non-Triviality and Object Proximity Non-Triviality demonstrations using a random action policy are collected. We assume that the random action policy performs poorly on the task. The specifications are checked to see if they are satisfied for $N=10$ demonstrations. Assuming the expert demonstrations complete the subtasks sequentially, an initial set of action masks can be obtained by recording the actions taken while each MBRM is running. These action masks are then provided to the LLM for further refinement, enabling it to leverage the expert-derived action masks as a prior rather than generating them from scratch.

Using expert and random demonstrations enables testing on in-distribution data without relying on $\mathcal{E}^{\text{sym}}$. However, such demonstrations are biased by the strategies of the expert and random policies, may not cover all possible behaviors, and do not provide theoretical guarantees. For instance, in \textit{LockedRoom}, an expert might always drop the key after opening a door before proceeding. As a result, all expert demonstrations would violate the Non-regressive Maximal Reward specification, resulting in the LLM being re-prompted. In reality, there exist demonstrations where the key is held until the agent reaches the goal, but the expert policy did not explore this strategy. Therefore, if one were to replace the SMT solver with demonstrations, multiple expert policies should be considered to improve coverage of potential behaviors.